\documentclass[letter]{ieice}
\usepackage[pdftex]{graphicx,xcolor}
\usepackage[fleqn]{amsmath}
\usepackage{newtxtext}
\usepackage[varg]{newtxmath}
\usepackage{verbatim}
\usepackage{float}
\usepackage{stfloats}
\usepackage{graphicx}
\usepackage{multirow}
\usepackage{multicol}
\usepackage{threeparttable}

\field{}
\title{Prior Information based Decomposition and Reconstruction Learning for Micro-Expression Recognition}
\titlenote{This work was partly supported by the Postgraduate Research and Practice Innovation Program of Jiangsu Province (Grant KYCX19$\_$0899), partly by the National Natural Science Foundation of China (NSFC) under Grants 72074038, 62076122 and 61971236.}
\authorlist{%
\authorentry{Jinsheng Wei}{n}{a}
\authorentry{Haoyu Chen}{n}{a}
\authorentry[lugm@njupt.edu.cn]{Guanming Lu}{n}{a}
\authorentry{Jingjie Yan}{n}{a}
\authorentry{Yue XIE}{m}{a}\MembershipNumber{2181943}
\authorentry{Guoying Zhao}{n}{a}
}

\affiliate[affiliate label]{Jinsheng Wei, Guanming Lu and Jingjie Yan are with College of Telecommunications and Information Engineering, Nanjing University of Posts and Telecommunications,
Nanjing 210003, China. Haoyu Chen and Guoying Zhao are with Center for Machine Vision and Signal Analysis, Oulu of University, 90014 Oulu, Finland. Yue Xie is with the School of Information and Communication Engineering, Nanjing Institute of Technology, Nanjing 211167, China.  
}
\paffiliate[present affiliate label]{Presently, the author is with the }

\begin{document}
\maketitle
\begin{summary}

Micro-expression recognition (MER) draws intensive research interest as micro-expressions (MEs) can infer genuine emotions. Prior information can guide the model to learn discriminative ME features effectively. However, most works focus on researching the general models with a stronger representation ability to adaptively aggregate ME movement information in a holistic way, which may ignore the prior information and properties of MEs. To solve this issue, driven by the prior information that the category of ME can be inferred by the relationship between the actions of facial different components, this work designs a novel model that can conform to this prior information and learn ME movement features in an interpretable way. Specifically, this paper proposes a Decomposition and Reconstruction-based Graph Representation Learning (DeRe-GRL) model to efectively learn high-level ME features. DeRe-GRL includes two modules: Action Decomposition Module (ADM) and Relation Reconstruction Module (RRM), where ADM learns action features of facial key components and RRM explores the relationship between these action features. Based on facial key components, ADM divides the geometric movement features extracted by the graph model-based backbone into several sub-features, and learns the map matrix to map these sub-features into multiple action features; then, RRM learns weights to weight all action features to build the relationship between action features. The experimental results demonstrate the effectiveness of the proposed modules, and the proposed method achieves competitive performance.

\end{summary}
\begin{keywords}
Micro-Expression Recognition; Decomposition and Reconstruction; Prior Information; Action Features.
\end{keywords}

\section{Introduction}

As a branch of affective computing, 
Micro-Expression Recognition (MER) is challenging due to the subtle facial muscle movements and fleeting duration. 
Micro-Expressions (MEs) can imply the true human emotions and help reveal the real psychological activities. Thus, MER has valuable potential applications \cite{takalkar2018survey,li2022deep}, such as medical diagnosis, emotion interfaces, security, and lie detection.

In recent years, deep learning methods are widely applied to MER, and present promising results \cite{li2022deep}. These methods employed deep models to extract ME features from ME videos or images. First, some works \cite{LEARNet-Verma,li2022deep} adopt the models with single stream structure to aggregate spatial or spatial-temporal information; second, in a lot of works, the models with two-stream or multi-stream structures \cite{li2022deep,3D-FCNN,STSTNet} are designed to aggregate more feature information from different views; finally, the models with cascade structure are also effective in MER, such as
Convolutional Neural Networks (CNN)+Graph Convolution Networks (GCN) \cite{AU-ICGAN}. Above works input ME video samples to the general models with single or combination structures to extract the ME features adaptively, which maybe not consider the prior information and peculiarities of MEs in terms of the design of models. Thus, to a certain extent, these models lack the interpretability and uniqueness for MER. 

To meet the prior information that a ME category can be inferred by the relationship between different facial component actions \cite{AU-ICGAN}, inspired by the decomposition and reconstruction ideas \cite{FDeReL} to decompose basic features into fine-grained features and then reconstruct  high-level features, this paper proposes a novel Decomposition and Reconstruction Graph Representation Learning (DeRe-GRL) model to introduce the prior information into the model, thereby enhancing the interpretability and feature learning ability of the model. DeRe-GRL  can enhance the geometric graph representation from facial landmarks and includes Action Decomposition Module (ADM) and Relation Reconstruction Module(RRM). Specifically, ADM can learns the different action features from from facial landmarks. RRM can learn a group of weights to build the relationship between different action features, thereby inferring ME categories. Furthermore, a set of losses are designed to constrain the action features and weights so that these weights can represent a reasonable relationship between different action features.

\section{Decomposition and Reconstruction-based Graph Representation Learning Method}

The proposed method takes facial landmarks as the input to aggregate the geometric movement information. First, based on the facial landmarks of the keyframes, a spatial-temporal graph is built. Second, a graph network is employed as the backbone network to extract basic features from the spatial-temporal graph. Third, according to facial three components including eyebrows, nose and mouth, ADM divides the basic features into three sub-features and maps three sub-features into different action features. Finally, RRM learns a group of weights to weight different action features and build their relationship to get ME features. The details are as follows.

\begin{figure*}
\setlength{\abovecaptionskip}{0pt}
\setlength{\belowcaptionskip}{-10pt}
\vspace{0pt}
\begin{center}
\includegraphics[width=0.9\linewidth]{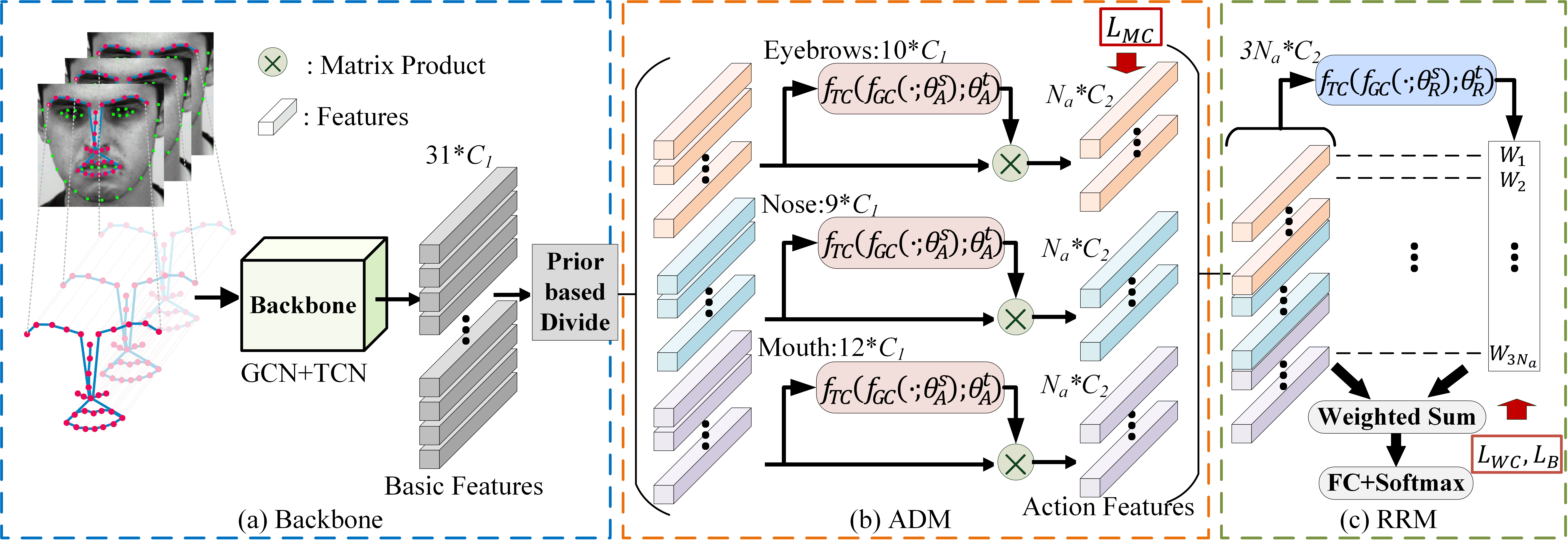}
\end{center}
   \caption{
   The flowchart of DeRe-GRL. The prior-based divide operation divides the basic features into three sub-features based on the prior information about the natural division of facial components. 
   }
\label{DeRe}
\end{figure*}


\subsection{Decomposition and Reconstruction-based Graph Representation Learning Model}


\leftline{\textbf{Spatial-Temporal Graph}}

In a ME video, the onset, apex, and offset frames can express the key dynamic process of ME occurrence. As following \cite{liu2019neural,G-TCN}, the three frames can be found from the annotations on datasets, and their detection belongs to another task \cite{Surveyspottingkh}. Thus, facial landmarks of the three keyframes are used to build a spatial-temporal graph as shown in Fig.\ref{DeRe} (a). Specifically, for every frame, 31 key landmarks are selected from the original 68 landmarks to cover the key information of eyebrows, nose and mouth (the landmarks in eye regions are not selected because there are noises in eye regions, such as quick blink \cite{HIGO-Li,G-TCN}.). For every frame, taking the 31 key landmarks as nodes, the spatial graph $G_0$ is built. By connecting the corresponding nodes between three keyframes, a spatial-temporal graph $G$ is built to model the geometric movement information of MEs.


The landmark coordinates include the geometric movement information that is discriminative for MER \cite{LFM}. Thus, this work adopts landmark coordinates as the node features of $G$, which is a very compact ME representation.

\leftline{\textbf{Graph Model-based Backbone}}

We construct a backbone that adopts GCN and Temporal Convolutional Network (TCN) to aggregate the spatial and temporal information in the designed graph, respectively. Concretely, for $i$-th layer, given input graph $G$ with node feature $X^i \in \mathbb{R}^{3*31*C^i}$ ($C^i$ is the dimension of node features), a spatial feature extractor $f_{GC}(X^i;\theta_i^s)$ is constructed to extract spatial features. Then, a temporal feature extractor $f_{TC}(f_{GC}(X^i;\theta_i^s);\theta_i^t)$ is construct to extract temporal features, where $\theta_i^s$ and $\theta_i^t$ are the trainable parameters. Thus, the constructed backbone can aggregate the spatial and temporal information to get basic features $X_B \in \mathbb{R}^{3*31*C_1}$, where $C_1$ is the dimension of output node features. Note that the temporal dimension is fixed to 3, so below, $X_B \in \mathbb{R}^{3*31*C_1}$ is denoted as $X_B \in \mathbb{R}^{31*C_1}$  to simplify the expression.

\leftline{\textbf{Action Decomposition Module}}


To aggregate the geometric movement information of different components individually, 31 nodes are divided into three parts corresponded to eyebrows, nose and mouth region. Thus, as shown in Fig.\ref{DeRe}, $X_B$ is divided into three sub-features: $X_B^e \in \mathbb{R}^{10*C_1}$, $X_B^n \in \mathbb{R}^{9*C_1}$ and $X_B^m \in \mathbb{R}^{12*C_1}$, and $G$ is correspondingly divided into: $G^e$, $G^n$ and $G^m$.

Next, every sub-feature is mapped into multiple action features based on one facial component. Supposed a sub-feature is expressed as $X_B^o \in \mathbb{R}^{N_f*C_1}$, where $N_f$ is the node number. Inspired by graph-based tensor grouping \cite{GHRD}, GC and TC are employed to find an embedding $ \mathbb{R}^{N_f*C_1} \rightarrow \mathbb{R}^{N_a*C_1}$ by forming the map matrix:
\begin{equation}
\begin{split}
M_{A}^{\prime}=f_{TC}(f_{GC}(X_B^o;\theta_{A}^s);\theta_{A}^t)\in \mathbb{R}^{N_f*N_a},
\end{split}
\label{AML}
\end{equation}
where $\theta_{A}^s$ and $\theta_{A}^t$ are the learnable parameters of GC and TC, respectively,  and $N_a$ is the number of the mapped action features. To make the $N_a$ action features are the probability combination of original $N_f$ features, $M_{A}$ is transfer to:
\begin{equation}
\begin{split}
M_{A-i,j}=\frac{e^{M_{A-i,j}^{\prime}}}{\sum_{j=1}^{N_a} e^{M_{A-i,j}^{\prime}}}\in \mathbb{R}^{N_f*N_a}.
\end{split}
\label{AML1}
\end{equation}
Then, the sub-feature $X_B^o$ is mapped into a group of action feature $F_A^o$ that includes $N_a$ sub-action features:
\begin{equation}
\begin{split}
F_A^o=M_A^T X_B^o \in \mathbb{R}^{N_a*C_1}.
\end{split}
\label{AF}
\end{equation}
Finally, concatenating three groups of action features obtains the final action features $F_A\in \mathbb{R}^{3N_a*C_1}$.

\leftline{\textbf{Relation Reconstruction Module}}
Inspired by the attention mechanism, RRM learns a group of weights to weight different action features for associating different action features to reconstruct high-level ME features. First, GC and TC are employed to deal with $F_A$:
\begin{equation}
\begin{split}
F^{\prime}=f_{TC}(f_{GC}(F_A;\theta_{R}^s);\theta_{R}^t) \in \mathbb{R}^{3N_a*C_2},
\end{split}
\label{W}
\end{equation}
where $\theta_{R}^s$ and $\theta_{R}^t$ are the learnable parameters, and $C_2$ is the dimension of node features. Then, the $L_1$ norm of $F^{\prime}$ is calculated as the weights of action features, that is:
\begin{equation}
\begin{split}
W= \sum_{j=1}^{C_2} \lVert F_{:,j}^{\prime} \lVert_1 \in \mathbb{R}^{3N_a*1},
\end{split}
\label{W}
\end{equation}
where $\lVert \cdot\lVert_1$ is $L_1$ norm, and $F_{:,j}^{\prime}$ is the $j$-th column of $F^{\prime}$ . Finally, all action features are weighted by the learned weight $W$ to get the final ME features, that is:
\begin{equation}
\begin{split}
F_{ME}= W^TF_A.
\end{split}
\label{W1}
\end{equation}
\subsection{The Constrain for Action Features and Weights}
To make each action feature more compact, decreasing the intra-class distance for each action feature is available. Thus, the mean center loss, inspired by center loss \cite{FDeReL}, is designed to constrain the action features.
\begin{equation}
\begin{split}
L_{MC}= \frac{1}{2N}\sum_{i=1}^{3N_a} \lVert F_A^i - c_i\lVert_2^2,
\end{split}
\label{LMC}
\end{equation}
 where $N$ and $F_A^i$ denote the number of samples in a mini-batch and the i-th row of $F_A$; $c_i$ is the mean of the $F_A^i$ in a mini-batch, as the center of i-th action feature.
 
 Furthermore, establishing a relationship between the weights of action features and ME categories is crucial to the rationality of weights. Thus, the center loss is employed to constrain weights, aiming at a more closed distribution between the weights corresponding to different samples of the same ME category. Supposed the ME category of the $i$-th sample is $l_i$, a weight center loss is defined as follows:
 \begin{equation}
\begin{split}
L_{WC}= \frac{1}{N}\sum_{i=1}^{N} \lVert W_i - W_{l_i}\lVert_2^2,
\end{split}
\label{LC}
\end{equation}
where $W_i$ and $W_{l_i}$ are the weight of i-th ME sample and the weight center of $l_i$-th ME category, respectively. However, in practice, a bit amount of action features (even only one) have significantly higher weights than other action features. To overcome this problem, introducing a balance loss \cite{FDeReL} :
 \begin{equation}
\begin{split}
L_{B}= \lVert\frac{1}{N}\sum_{i=1}^{N}  W_i - W_m\lVert_2^2,
\end{split}
\label{LC}
\end{equation}
where $W_m=[\frac{1}{3N_a},...,\frac{1}{3N_a}] \in \mathbb{R}^{3N_a*1}$. So, the total loss is:
 \begin{equation}
\begin{split}
L= L_{ME}+\lambda_1L_{MC}+\lambda_2L_{WC}+\lambda_3L_{B},
\end{split}
\label{LC}
\end{equation}
where $L_{ME}$ is the  cross entropy loss for ME classification, and $\lambda_1$, $\lambda_2$ and $\lambda_3$ are trade-off parameters.

\section{Experiments}

The evaluation experiments are carried out on CASME \uppercase\expandafter{\romannumeral2} \cite{CASME2-Yan} and SAMM \cite{A12} datasets. CASME \uppercase\expandafter{\romannumeral2} includes 255 ME videos with 26 subjects; SAMM includes 159 ME videos with 32 subjects. Following \cite{LBP-FIP,G-TCN}, 247 and 136 ME videos with five categories of MEs are selected on CASME \uppercase\expandafter{\romannumeral2} and SAMM, respectively. In the experiment, the accuracy (Acc) and F1-score (F1) under leave-one-subject-out (LOSO) cross-validation are reported. The backbone has 2 layers. Also, the node feature dimension $C_1$ of the backbone output is set to 16 on CASME \uppercase\expandafter{\romannumeral2}. Since SAMM has a smaller sample size, $C_1$ is set to 8 on SAMM to alleviate overfitting. The optimal values are explored from [2,3,4,..., 8] for $N_a$ and [0.0001,0.001,0.1,1] for $\lambda_1$, $\lambda_2$ and $\lambda_3$. Based on the explored results, $N_a$, $\lambda_1$, $\lambda_2$ and $\lambda_3$ are empirically set to 3, 1, 1, and 0.1, respectively. Facial movements are magnified \cite{LVMM}, and the amplification factor is set to 3, thereby making the movements of landmarks more apparent. Dlib is employed to detect facial landmarks. Because the cropped face images with different sizes have different landmarks, we regard this difference as noise to augmented data.


\leftline{\textbf{Ablation Study}}

\begin{table}[t!]
\setlength{\abovecaptionskip}{0pt}
\setlength{\belowcaptionskip}{0pt}
\vspace{0pt}
\caption{The evaluation on ADM and RRM. Acc(\%).}
\begin{center}
\footnotesize
\begin{tabular}{|l|c|c|c|c|}
\hline
\multirow{2}{*}{Methods} &\multicolumn{2}{c|}{CAMSE\uppercase\expandafter{\romannumeral2}} &\multicolumn{2}{c|}{SAMM} \\
\cline{2-5}
&Acc  &F1 &Acc  &F1\\
\hline\hline
backbone  & 73.68 & 0.726  &66.18&0.640  \\
backbone + ADM  & 76.11& 0.758  &69.85&0.688\\
backbone + ADM + RRM  & \textbf{80.16}& \textbf{0.796} &\textbf{72.06} &\textbf{0.699}\\
\hline
\end{tabular}
\end{center}
\label{AST}
\vspace{-5mm}
\end{table}
\begin{table}[t!]
\large
\setlength{\abovecaptionskip}{0pt}
\setlength{\belowcaptionskip}{0pt}
\caption{The evaluation on $L_{MC}$, $L_{WC}$ and $L_{B}$. Acc(\%).}
\begin{center}
\footnotesize
\begin{tabular}{|l|p{20pt}|c|c|p{20pt}|c|}
\hline
\multirow{2}{*}{Methods} &\multicolumn{2}{c|}{CAMSE\uppercase\expandafter{\romannumeral2}} &\multicolumn{2}{c|}{SAMM} \\
\cline{2-5}
&Acc  &F1 &Acc  &F1\\
\hline\hline
$L_{MC}+L_{WC}+L_{B}$  &\textbf{ 80.16}& \textbf{0.796} &\textbf{72.06}&\textbf{0.699}\\
$L_{WC}+L_{B}$($\lambda_1$=0) & 74.09& 0.730& 63.97& 0.582\\
$L_{MC}+L_{B}$($\lambda_2$=0) & 76.52& 0.762& 68.38& 0.646\\
$L_{MC}+L_{WC}$($\lambda_3$=0) & 76.11& 0.759& 70.59& 0.684\\
\hline
\end{tabular}
\end{center}
\label{ASL}
\vspace{-6mm}
\end{table}
First, ADM and RRM are evaluated as shown in Table \ref{AST}. Compared to the backbone, the performance is improved after adopting ADM, which demonstrates the effectiveness of ADM. That is, decomposing the facial features and learning the action features of facial different components separately are beneficial to representing ME features. Furthermore, DeRe-GRL (backbone+ADM+RRM) gets the best performance, which shows the effectiveness of RRM and DeRe-GRL. That is, RRM can effectively build the relationship of different action features, and the model structure of decomposition and reconstruction based on prior information can enhance the geometric movement features effectively.

$L_{MC}$, $L_{WC}$ and $L_{B}$ are evaluated by removing a single loss in turn. Table \ref{ASL} shows the evaluated results. As a whole, the performance degrades no matter which loss is removed, thereby demonstrating the positive effects of the three losses. In fact, the three losses are crucial to extracting the compact action features and learning the weights that can build the relationship between these action features. Interestingly, among the three losses, $L_{MC}$ has a more important impact, which demonstrates that DeRe-GRL has higher requirements on the intra-class distance of action features.

The trade-off parameters ( $\lambda_1$, $\lambda_2$ and $\lambda_3$) of three losses are set to the optimal values: 1, 1, and 0.1, respectively. It turns out that 1) the relatively large trade-off parameters for $L_{MC}$ and $L_{WC}$ are beneficial to improve performance. Namely, the action features and weights with a more closed attribution can effectively reconstruct high-level ME features. 2) A relatively small $\lambda_3$ is suitable for $L_{B}$ to prevent overbalancing of weights that may lead to an unreasonable representation of the relationship between action features.

\leftline{\textbf{Comparing with other methods}}

As shown in Table \ref{CO}, the proposed method is compared with other methods. In traditional methods, KTGSL is state-of-the-art. Compared to it, DeRe-GRL gets better performance on both datasets. In deep learning methods, LGCcon and G-TCN are state-of-the-art methods. Compared to LGCcon, DeRe-GRL has an obvious advantage. Furthermore, compared to G-TCN, our method still maintains an obvious advantage on CASME \uppercase\expandafter{\romannumeral2}. However, on SAMM, G-TCN has better accuracy than DeRe-GRL. The reason may be that compared to G-TCN, DeRe-GRL has more learnable parameters, which causes DeRe-GRL to be more prone to overfitting on SAMM with a smaller sample size. LFM also adopts facial landmarks as input. Compared to LFM, DeRe-GRL has a better performance, which demonstrates DeRe-GRL can effectively aggregate discriminative geometric movement information from facial landmarks for MER. 

Overall, DeRe-GRL achieves better performance, compared to the most existing methods. This advantage mainly is derived from that 1) the magnified movement of facial landmarks contains rich discriminative information for MER; 2), the proposed method considers the prior information to guide the model to effectively learn high-level ME features.


\begin{table}[t!]
\large
\setlength{\abovecaptionskip}{0pt}
\setlength{\belowcaptionskip}{0pt}
\vspace{0pt}
\caption{Comparing with other methods.  Acc(\%).}
\begin{center}
\footnotesize
\begin{tabular}{|l|p{20pt}|c|c|p{20pt}|c|}
\hline
\multirow{2}{*}{Methods} &\multicolumn{2}{c|}{CAMSE \uppercase\expandafter{\romannumeral2}} &\multicolumn{2}{c|}{SAMM} \\
\cline{2-5}
&Acc  &F1 &Acc  &F1\\
\hline\hline
LBP-SDG~\cite{Wei}&71.32& 0.665&N/A& N/A\\
DSSN \cite{DSSN-Khor}& 70.78& 0.729& 57.35& 0.464\\
LFM \cite{LFM} &73.98&0.717&N/A&N/A\\
LGCcon \cite{LGCcon-Yante} & 65.02& 0.640& 40.90& 0.340\\
G-TCN \cite{G-TCN} & 73.98& 0.725& \textbf{75.00}& 0.699\\
KTGSL~\cite{Wei-KTGSL}&72.58& 0.682&56.11& 0.493\\

DeRe-GRL & \textbf{80.16}& \textbf{0.796} &72.06&\textbf{0.699}\\
\hline
\end{tabular}
\end{center}
\label{CO}
\vspace{-5mm}
\end{table}

\section{Conclusion}

By introducing the prior information to guide the model to learn high-level ME features, this paper proposes DeRe-GRL model to extract action features and build the relationship between different facial actions. DeRe-GRL includes two key modules: ADM and RRM. The experimental results show that ADM can effectively extract the action features by decomposing different facial components and learning a map matrix, and RRM can learn the weights of different action features to reconstruct ME features. Furthermore, the designed three losses are crucial to learning different action features and weights. The evaluation experiments show that every loss has an important impact and the intra-class distance of action features is more crucial to the performance of DeRe-GRL. Overall, the proposed method can enhance the geometric movement features from facial landmarks in an interpretable way and achieves competitive results.

\bibliographystyle{ieicetr}
\bibliography{mybibfile}


\end{document}